# ANEC: An Amharic Named Entity Corpus and Transformer Based Recognizer

**EBRAHIM CHEKOL JIBRIL AND A. CÜNEYD TANTUĞ**

Faculty of Computer Engineering and Informatics, Istanbul Technical University, 34469 Istanbul, Turkey

Corresponding author: Ebrahim Chekol Jibril (jibril@itu.edu.tr)

The work of Ebrahim Chekol Jibril was supported by the Turkcell-Istanbul Technical University Researcher Funding Program.

**ABSTRACT** Named Entity Recognition is an information extraction task that serves as a preprocessing step for other natural language processing tasks, such as machine translation, information retrieval, and question answering. Named entity recognition enables the identification of proper names as well as temporal and numeric expressions in an open domain text. For Semitic languages such as Arabic, Amharic, and Hebrew, the named entity recognition task is more challenging due to the heavily inflected structure of these languages. In this paper, we present an Amharic named entity recognition system based on bidirectional long short-term memory with a conditional random fields layer. We annotate a new Amharic named entity recognition dataset (8,070 sentences, which has 182,691 tokens) and apply Synthetic Minority Over-sampling Technique to our dataset to mitigate the imbalanced classification problem. Our named entity recognition system achieves an F_1 score of 93%, which is the new state-of-the-art result for Amharic named entity recognition. We will release the dataset after acceptance.

**INDEX TERMS** Named Entity Recognition. Synthetic Minority Over-sampling Technique. Deep Neural Network. BiLSTM-CRF.

## ACRONYMS

| | |
|---|---|
| ANERS | Amharic Named Entity Recognition System |
| BiLSTM | Bidirectional Long Short-Term Memory |
| BiLSTM-CRF | Bidirectional Long Short-Term Memory-Conditional Random Fields |
| CRF | Conditional Random Fields |
| HMM | Hidden Markov Models |
| IOB | Inside-Outside-Beginning |
| LOC | Location |
| LSTM | Long Short-Term Memory |
| NER | Named Entity Recognition System |
| NLP | Natural Language Processing |
| ORG | Organization |
| PER | Person |
| POS | Part-of-Speech |
| QA | Question answering |
| RNN | Recurrent Neural Networks |
| SVM | Support Vector Machine |
| SMOTE | Synthetic Minority Over-sampling Technique |
| SERA | System for Ethiopic Representation in ASCII |
| TTL | Title |

## I. INTRODUCTION

Named Entity Recognition (NER) is a subtask of information extraction, which extracts and classifies specific predefined types of entities, which may be proper names, numerical, and temporal expressions. Usually, person, location and organization names are considered as proper names in most of the studies. Numerical expressions usually cover numeral, money, and percentage expressions, while date and time expressions are classified as temporal expressions [29]. The NER task is a non-trivial task since using simple lookup lists to capture these expressions in the running text is not sufficient due to the infinite number of elements in these types. It is not feasible to construct a large set of people names in which all the names that can be seen in a text is contained [70]. The same fact is also valid for the location and organization names. For these Named Entity (NE) types, we need a smarter way of identifying NEs according to the context, mostly by Machine Learning (ML) techniques. On the other hand, rule-based systems generally perform quite well in the identification of temporal and numerical expressions.

A NER system is usually a pre-processing step within large Natural Language Processing systems. The quality of the NER system has a direct impact on the performance of the overall NLP system. Many research efforts have been conducted to prove the importance of NER to the other Natural Language Processing tasks such as Question answering (QA) [28]. The majority of the considered questions expect a named entity or a list of named entities as answers. Greenwood and Gaizauskas [28] use a NER system to improve the performance of the answer extraction module based on a pattern matching approach. The authors use the NER system to capture the answers, which are not possible to capture using only patterns. They report that the NER system improved the accuracy of answering the questions.

In search result clustering, the use of a NER system before comparing the contents of the documents proved to be very useful [66]. In machine translation, Babych, B. and A. Hartley [5] proved using NER systems as a pre-processing task improves the quality of the translation output.

Traditional sequence labeling models are linear statistical models, including Hidden Markov Models (HMM) and Conditional Random Fields (CRF) [55][49][39], which rely heavily on hand-crafted features and task-specific resources. For example, English NER benefits from carefully designed word spelling features; orthographic features, and external resources such as gazetteers. However, such task-specific knowledge is usually costly to develop, making sequence labeling models difficult to adapt to new tasks or new domains.

In the past years, non-linear neural networks with word embeddings have been broadly applied to NLP problems with great success. Collobert et al. [12] proposed a simple but effective feed-forward neural network that independently classifies labels for each word by using contexts within a fixed-size window. Recurrent Neural Networks (RNN) together with its variants such as Long-Short Term Memory (LSTM) [31] have shown great success in modelling sequential data. Several RNN-based neural network models have been proposed to solve sequence labeling tasks like speech recognition [27], POS tagging [33], and NER [33] [11]. Transformer is one of the most commonly used neural network architectures in natural language processing. The transformer architecture consists of stacked transformer layers, each of which takes a sequence of vectors as input and outputs a new sequence of vectors with the same shape [67]. Recently, transformer-based architectures are shown to achieve better performance against traditional models for NER and Question Answering tasks [16] [38].

Most studies conducted for widely spoken languages such as English, German, Chinese and Arabic. However, there exists a limited number of studies on NER for low-resource languages. In this work, we aim to build a large dataset by manually annotating entities and build an Amharic Named Entity Recognition system by using deep neural networks.

There is limited research study, which was carried out to prove the efficiency of the BiLSTM-CRF model for NER in Amharic texts. Therefore, our main contributions in this study are building a large publicly available Amharic NER dataset and a NER architecture achieving the state-of-the-art results for the Amharic NER task. The rest of the paper is structured as follows. In the second section of this paper, we will give brief information about Amharic and an overview of the Amharic language peculiarities. In section three we describe previous works related to the Amharic NER task. Section four is dedicated to giving a brief description of our model. Details about the evaluation data we used in our experiments are given in Section five. The results of our experiments with BiLSTM-CRF and a comparison with previous work results are presented in Section six. Finally, in Section seven, we draw some conclusions and discuss future works.

## II. THE AMHARIC LANGUAGE

Amharic is a member of Semitic language family such as Arabic, Syriac, and Hebrew [48]. The language is spoken by more than 50 million people as their mother language and over 100 million as a second language in Ethiopia [44] [64]. Amharic is the second most spoken Semitic language after Arabic [20]. It is also the second most spoken language in Ethiopia and one of the five widely spoken languages on the African continent. The Amharic language is mainly spoken in Ethiopia and Eritrea. It is also the working language of the federal government of Ethiopia and also non-government organizations as well as private institutions.

The writing system of Amharic is an abugida, which has originated from the Ge'ez scripts known as ፊደል (Fidäl), which is now only used in the Ethiopian Orthodox Church. Amharic language writing system is composed of a total of 238 characters among which are combinations of the 33 core characters and special character (ñ) ("V"). Each of these core characters and special character occurs in 7 forms (orders); one basic form and six non–basic forms representing syllable combinations consisting of a consonant and following vowel. The non-basic forms are derived from the basic forms by generally regular modifications (see Figure 1). For example, the second-order characters are formed mostly by attaching strokes to the right of the character [4].

| | |
|---|---|
| ሀ(hä) | ሁ(hu) |
| መ(mä) | ሙ(mu) |
| ሰ(sä) | ሱ(su) |

**Figure 1: Character formation**

In addition to the set of these 238 characters, there are 50 labialized characters, 9 punctuation marks and 20 numerals. These bring the total number of characters in the script to 317 [6] [7].

Not all the letters of the Amharic script are strictly necessary for the pronunciation patterns of the spoken language; some were simply inherited from Ge'ez without having any semantic or phonetic distinction in modern Amharic. Most of the labialized consonants, which are simply inherited from Ge'ez, are redundant. The language also has its own unique set of punctuation marks ("፥/word separator", "።/full stop", "፣/comma", "፤/colon", "፤/semi-colon", "፦/preface-colon", "፧/question mark"( It is not used anymore), "※/section marker", "፨/ paragraph separator"). Unlike other Semitic languages, Amharic is written from left to right.

## The Challenges of NER in the Amharic Language

Conducting a NER study in Amharic necessitates dealing with some challenges, mostly due to its rich morphological structure and unique orthography [8].

### Lack of Capitalization

In Amharic language, there is no capitalization when writing proper nouns. As in English and most European languages, proper nouns are written by making the initial letter of the word in uppercase, which provides strong evidence for the identification and classification of named entities (NE).

From a general viewpoint, the NER task can be considered as a composition of two sub-tasks: First, the detection of the existing NEs in a text, which is an easy sub-task if we can use the capital letters as indicators to determine where the NEs start and end. However, this is trivial only when the target language supports capital letters, which is not the case for the Semitic language family (Amharic, Arabic and others). *Figure 2* shows the example of two words where only one of them is a NE but both of them start with the same character (example sentence along with its English translation). In **example sentence1** ሰላም/säläm/ is a named entity whereas, in **example sentence2** it is not. The absence of capital letters in the Amharic language seems to be a major obstacle to obtain high performance in Amharic NER. The same problem is also exhibited even in languages that support capital letters for social media and informal texts where capitalization is generally ignored [24].

| | Example Sentence 1 | Example Sentence 2 |
|---|---|---|
| Amharic: | ሰላም ወደ ዩኒቨርሲቲ ሄደች | ሰላም ለሃገር ግንባታ አስፈላጊ ነው |
| English: | Selam went to university. | Peace is essential for nation building. |
| Tokens: | (1) ሰላም/säläm/ | (1) ሰላም/säläm |
| | (2) ወደ/wädä | (2) ለሃገር/lähägär |
| | (3) ዩኒቨርሲቲ/yunivärsiti | (3) ግንባታ/gbbata |
| | (4) ሄደች/hedäc | (4) አስፈላጊ/aäsfälagi |
| | | (5) ነው/näw |

**Figure 2**: An example illustrating the absence of capital letters in Amharic

### The Agglutinative Morphology

The Amharic language, like Arabic, has a highly agglutinative morphology in which a word may have prefixes, lemma and suffixes in different combinations resulting in a very complicated morphology. For instance, a person name like "መሉ አለም" (Mulu Alem) can be used as a location name when the morpheme "ħ" (kä) is put in front of it.

Similar to Arabic, Turkish and other languages, Amharic is a highly inflectional language, where a surface word is constructed with prefix(es), a lemma, and suffix(es), though prefix(es) and suffix(es) are optional. A prefix can be an article, a preposition or conjunction, whereas a suffix is generally an object or a personal/possessive anaphora. Both prefixes and suffixes are allowed to be combinations, and thus a word can have zero or more affixes. Compared to texts written in other languages whose morphologies are not complex, this inflectional and derivational characteristic of the language makes Amharic texts sparser and thus most of the Amharic NLP tasks are harder and more challenging. However, concerning the classification subtask of NER, we can say that the classification of NEs relies mainly on the word-form and the context in which it appeared in the text in order to decide the class it belongs to.

### Orthographic Variation

Sometimes, an Amharic word may have different orthographies with the same pronunciation, still referring to the same word. For example, "አመት/amat -year" and "ዓመት/ämat - year" and also "መስጊድ/mäsgid -mosque" and "መሥጊድ/mäśgid -mosque" have the same meaning and pronunciation, but spelled differently. Orthographic variation increases the number of out of vocabulary (OOV) words which deteriorates the quality of word embeddings. In addition to this, it increases the number of unseen data in the training corpus.

## III. Related Works

In this section, we present recent and important studies on NER with focus on works that concentrate on Semitic languages. Most of the NER studies on Semitic languages are mainly in Arabic and partly in Hebrew. Early studies on Arabic NER are mainly based on handcrafted rules [41] [60]. Machine learning techniques, however, have played an important role in moving NER research forward by providing different learning methods such as Hidden Markov Model, Conditional Random Fields, Maximum

Entropy, Support Vector Machines and Deep Neural Network.

Most of the current named entity recognition research utilizes machine learning approaches that mainly depend on the availability of large scale training data [33] [11] [40], which is usually very hard to access for low-resource languages. In particular, most NER efforts have focused on a few European and Asian languages, while African languages have been given little attention. In literature, only four studies of NER on Amharic are encountered [43] [3] [8] [14]. In these Amharic NER studies, two NER datasets compiled from different sub-sets of Walta Information Center Corpus [13] are used. In addition to Walta Information Center corpus there are also Adelani [1] dataset and Sikdar and Gambäck [61] New Mexico State University Computing Research Laboratory dataset, which is annotated for the SAY project. The data is annotated with 6 classes (PER, LOC, ORG, TIME, TTL, and O-other) and it is available in GitHub[1]. Sikdar and Gambäck [61] employed a stack-based deep learning approach incorporating various semantic information sources that are built using an unsupervised learning algorithm with word2Vec, feature vector and one-hot vector extracted by using a CRF classifier. They have reported that the stack-based approach outperformed other deep learning algorithms. Sintayehu, and Lehal [62] applied graph-based label propagation algorithm for 6 classes (PER, LOC, ORG, DATE, MONEY, and O-other). The researchers compared expectation maximization with semi-supervised learning approaches. The experiment reveals that label propagation based NER achieves superior performance compared to expected maximization using a few labeled training data. Table 1 presents the methods and performances of the previous Amharic NER studies.

All of the aforementioned Amharic NER studies extracted subsets of randomly selected sentences containing at least one named entity from the Walta Information Center Corpus and these extracted sentences are manually annotated with 4-class named entity tags for persons, organizations, locations and others (non-NE). As a result, each study trains the machine learning models on a different training dataset and evaluates their performance on a separate dataset. Consequently, the lack of a standard benchmark test dataset limits a comparable evaluation of these previous studies.

---

[1] https://github.com/geezorg/data/tree/master/amharic/dictionaries/nmsu-say/

TABLE 1: PREVIOUS WORKS ON AMHARIC NER

| Author | Method | Corpus | Entities | F_1 Score |
|--------|--------|--------|----------|-----------|
| Mehamed [43] | Conditional Random Field | Custom Corpus (18,299 tokens) | Person, Organization, Location | 75.0% |
| Alemu [3] | Conditional Random Field | Alemu[3] data(13,538 token) | Person, Organization, Location | 80.7% |
| Belay [8] | Hybrid(J48 and SVM) | Alemu[3] Data(13,538 token) | Person, Organization, Location | 96.1% J48, 85.9 SVM |
| Demissie [14] | BiLSTM | Alemu[3] Data(13,538 token) | Person, Organization, Location | 92.6% |
| Sintayehu, and Lehal [62] | Semi-Supervised | Custom Corpus (109,676) | Person, Organization, Location, Date, Money | 79% |
| Sikdar and Gambäck [61] | Stack-based LSTM | SAY Corpus (109,676 token) | Person, Organization, Location, Time, Title | 74.26% |
| Adelani [1] | XLM-R-base | Own corpus data(39,531 tokens) | Person. Organization, Location, Date | 70.96% |

Both Mehamed [43] and Alemu [3] used Conditional Random Fields [35] classifiers trained on different word and context features (word prefixes and suffixes, and the NE and part-of speech tags of the word), using about 90% of their data for training and the remaining 10% for testing (without cross-validation). Mehamed [43] achieved recall, precision and F_1 score values of 75.0%, 74.2%, and 74.6%, respectively, for the NER task on a Walta Information Center corpus subset consisting of 10,405 words, of which 961 were used for testing. Within the test data only 96 named entities exist.

Alemu [3] stated that the data used by [43] is not accessible, so the researcher experimented on another 13,538-word subset of the Walta Information Center corpus, of which 1,242 words were used for testing. Using context windows of up to two words before and after the current token, Alemu [3] achieved a recall of 84.9% and precision 76.8% and 80.7% to F_1 score. The small size of the test datasets makes the evaluations in both researches unreliable, as stated by the authors in their publications. Mehamed [43] found that part-of-speech tagging improved the results while using word prefixes did not. Alemu [3] on the other hand claimed that word prefixes contributed positively, while part-of-speech tagging did not. However, they both agreed that context and word suffixes were useful features.

Belay [8] included all features previously used and utilized a combination of decision trees, support vector machines and hand-crafted rules on the dataset created by [3], and artificially expanded it to get a better balance between the classes by using Synthetic Minority Over-sampling Technique (SMOTE) technique. They use two rules which classify words into NE types based on the occurrence of trigger words in a given sentence. Using a list of trigger words that appear before and after NEs: Searching at the beginning of the sentence like Personal names and Searching at the end of the sentence like organization names. To balance the datasets, this work expanded the other classes (person, location and organization) so that the actual training dataset contained 31,347 tokens and then apparently tested on 40% of the same data as used for training. Demissie [14] used word vectors as features instead of manually designed features and tested the automatically extracted word vector features with different classifiers (SVM, J48, LSTM and BiLSTM). To generate the word vectors, the WIC website archive, which contains 611,294 untagged tokens, is used. Demissie [14] trained the classifier with the same datasets as [8]. Both Belay [8] and Demissie [14] applied SMOTE to balance between classes. Demissie [14] split the datasets into 80% for training and 20% for testing purposes. Their experiments showed that word vector features could substitute manually designed features while maintaining high performance for Amharic NER. Compared to the SVM, the score obtained by BiLSTM is 2.9% lower. The justification for these results is the small number of datasets used for BiLSTM (deep neural network requires large datasets), testing datasets used for BiLSTM are two-fold of SVM and the last reason was the parameters used for the network may not be optimized properly.

Recently it is common to use a hybrid NER architecture by using two or more different algorithms like rule based and SVM [8], rule based and decision tree (Belay, 2014), LSTM and CRF [33] [18] [47] [68]. These models achieved state-of-the-art for different languages like English, Arabic, Turkish, etc. However, such kind of a hybrid model is not tested for Amharic.

## 1. IMBALANCED DATASETS

The predictive classifiers generally suffer from imbalanced training datasets where the number of examples in the dataset for each class label is not balanced.

Imbalance on the order of 102 to 1 is prevalent in fraud detection and an imbalance of up to 105 to 1 has been reported in other applications [51]. In some real-world classification tasks like above, the unusual or interesting class is rare among the general population, the class distribution is imbalanced [34] [22]. An imbalanced training dataset is challenging for predictive classifiers since the model cannot be trained on a proper number of examples to learn the discriminative characteristics of the examples in the minority class. A feed-forward neural network trained on an imbalanced dataset may not learn to discriminate enough between classes [54].

There have been attempts to deal with imbalanced datasets in domains such as fraudulent telephone calls, telecommunications management, text classification and detection of oil spills in satellite images. Many researchers have addressed the issue of class imbalance in two ways. One is to assign distinct costs to training examples and the other is to re-sample the original dataset, either by under-sampling the majority class and/or oversampling the minority class. The problems of unequal error costs and uneven class distributions are related. It has been suggested that, for training, high-cost instances can be compensated by increasing the dataset [21].

A naive approach of oversampling is duplicating examples in the minority class. However, these additional examples are not capable of introducing new information to the model. As an alternative oversampling approach, new examples can be synthesized from the existing examples. This type of data augmentation for the minority class is named as Synthetic Minority Oversampling Technique (SMOTE). SMOTE is a means of increasing the sensitivity of a classifier to the minority class by balancing the number of instances between the classes. SMOTE generate synthetic examples by operating in feature space rather than data space. The minority class is over-sampled by taking each minority class sample and introducing synthetic examples using nearest neighbors algorithms. Depending upon the amount of over-sampling required, neighbors from the k nearest neighbors are randomly chosen. The SMOTE implementation used five nearest neighbors. Synthetic samples are generated in the following way: Take the difference between the feature vector (sample) under consideration and its nearest neighbor. Multiply this difference by a random number between 0 and 1, and add it to the feature vector under consideration. This causes the selection of a random point along the line segment between two specific features. This approach effectively forces the decision region of the minority class to become more general [9]. The algorithm of SMOTE taken from [9] is presented in Appendix I.

## IV.Model
In this section, we describe the architecture of our Amharic NER model. We first define the input/output representation. Next, we provide information on LSTM and Bi-directional LSTM. Finally, we explain how the CRF-based decoder model is built.

### 1. INPUT REPRESENTATION

#### A. CHARACTER EMBEDDING
Sub-word units play a substantial role in word representations, peculiarly in languages with complex morphological structures. Character-level representations have been found useful for morphologically rich languages and to handle the out-of-vocabulary problem. Learning character-level embeddings has the advantage of learning representations specific to the task and domain [36]. Extracting the prefix and suffix information of a surface requires a morphological analysis step. Embeddings are used instead of hand-engineering prefix and suffix information about words. Instead, we opt for relying on the word vector representation of the word formed by a BiLSTM which accepts the characters of the word as input. A randomly initialized character look-up table contains an embedding for every character. The character embeddings belonging to every character in a word is processed both left-to-right and right-to-left directions by the BiLSTM Layer (Figure 3). The final character-level representation of a word is the concatenation of its forward and backward representations from the BiLSTM Layer. This character-level representation is also concatenated with a word-level representation which is described in the next subsection.

#### B. WORD EMBEDDING

Word Embeddings are vector representations of words that allow words with similar meanings to be represented as vectors close to each other in the vector space [46]. Since learning too many parameters from limited data is difficult, obtaining word embeddings on a large unlabeled corpus is a widely used technique. Instead of randomly initializing embeddings, using pre-trained embeddings have improved the performance of neural network architecture [36]. Like the work of Demissie [14], we used pre-trained word embeddings for our work and observed significant improvements. Embeddings are trained using the Continuous Bag of Words (CBOW) model.

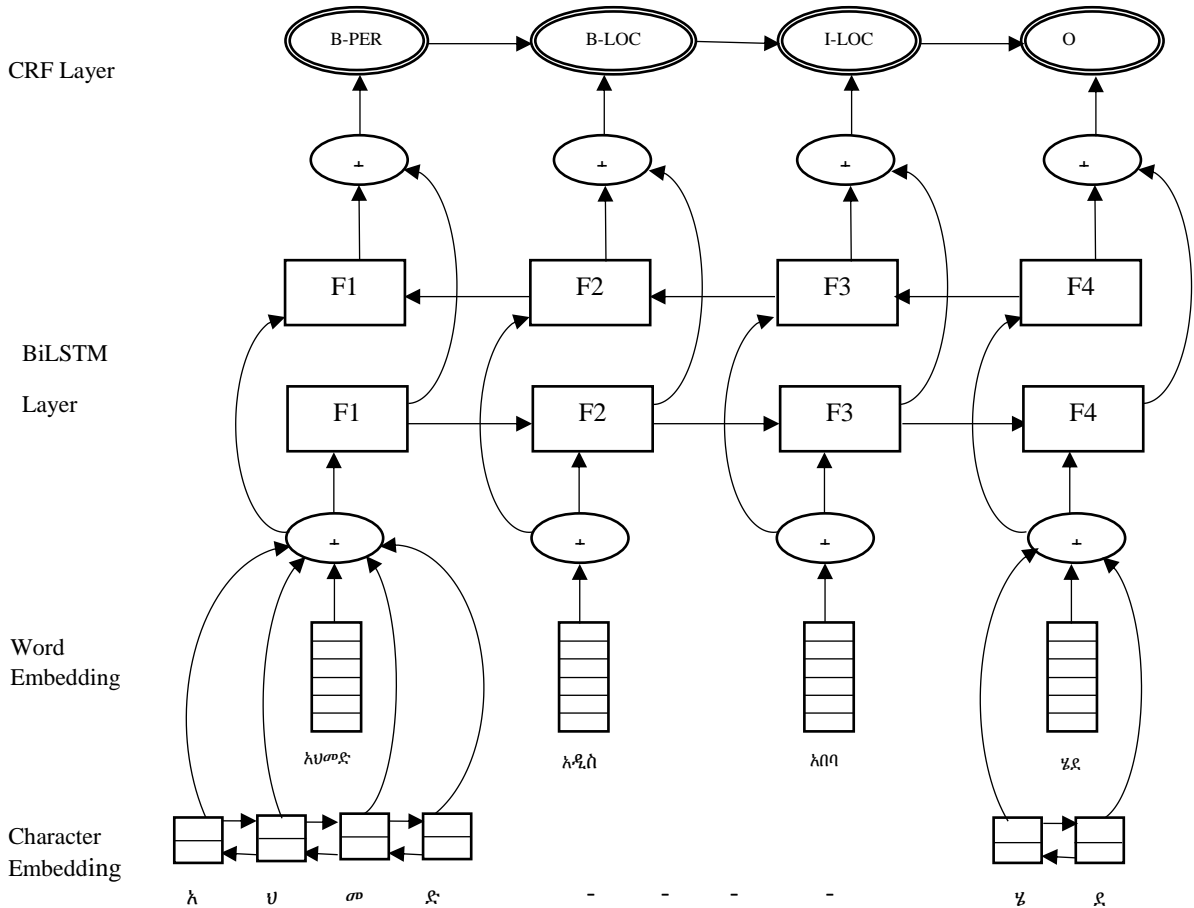

We used a pre-trained fastText[2] Amharic word embeddings model with an embedding dimension of 300 and a window size of five.

---

[2] https://fasttext.cc/docs/en/crawl-vectors.html

### 1. LSTM

Recurrent Neural Networks are shown to be able to achieve high performances in many natural language processing tasks such as language modelling [45], parsing [19], and machine translation [63]. One major problem with simple RNNs is that they are difficult to train for long distance dependencies due to the vanishing and the exploding gradient problems. Hochreiter and Schmidhuber [31] proposed Long Short-Term Memory to overcome the long-term dependency problem. They introduced a special memory cell, which is controlled by input, output and forget gates. The input gate controls how much new information should be added to the current cell, the forget gate controls what old information should be deleted. The output gate controls the information flow from the cell to the output. Long Short-Term Memory has emerged as an effective and scalable model for several learning problems related to sequential data. LSTMs are general and effective at capturing long term temporal dependencies. They do not suffer from the optimization difficulty that plague simple recurrent networks [30] and have been used to advance the state-of-the-art for many difficult problems.

The LSTM memory cell is defined by the following equations:

$$f_t = \sigma\left(W_{fx}x_t + W_{fh}h_{t-1} + w_{fc}c_{t-1} + b_f\right) \tag{1}$$

$$i_t = \sigma\left(W_{ix}x_t + W_{ih}h_{t-1} + w_{ic}c_{t-1} + b_i\right) \tag{2}$$

$$c_t = f_t \odot c_{t-1} + i_t \odot \tan h\left(W_{ix}x_t + W_{ch}h_{t-1} + b_c\right) \tag{3}$$

$$o_t = \sigma\left(W_{ox}x_t + W_{oh}h_{t-1} + w_{oc}c_t + b_o\right) \tag{4}$$

$$h_t = o_t \odot tanh(c_t) \tag{5}$$

Where $\sigma$ is element-wise logistic sigmoid function, *tanh* is the hyperbolic tangent function and $\odot$ is element-wise product, $W$'s are weight matrices, and $b$'s are biases. $f$, $I$ and $o$ are the forget, input and output gates respectively, $c$ denotes the cell vector, and $h$ is the hidden state vector. All gate vectors and the cell vector have the same dimensionality as the hidden state vector.

It is a common approach to use both preceding and following tokens to derive features for the current tokens in natural language processing tasks. When we look at the LSTM equations, the current output depends only on previous inputs, the initial cell value and hidden state.

Graves and Schmidhuber [26] proposed BiLSTM to gain information from future inputs. In a BiLSTM, two LSTM components are present, namely the forward LSTM and backward LSTM. The forward LSTM traverses the sequence in the forward direction and the backward LSTM traverses the same sequence in the reverse order using $h_{t+1}$ and $c_{t+1}$ are used instead of $h_{t-1}$ and $c_{t-1}$ for the gate calculations. In a bidirectional model the output at time $t$ depends on both the forward hidden state $\overrightarrow{h_t}$ and the backward hidden state $h_t$. The output of character and word embeddings are concatenated as an input to the BiLSTM.

### 2. CRF (CONDITIONAL RANDOM FIELD)

For sequence labeling tasks, it is important to consider the correlations between labels in neighbourhoods and jointly decode the best sequence of output labels for a given input sentence. For example, in POS tagging an adjective is more likely to be followed by a noun than a verb, and in NER with standard IOB2 annotation I-LOC cannot follow I-ORG. The IOB2 annotation used a B-tag for all base noun phrase initial words [58]. Therefore, instead of decoding each label independently, we model label sequence jointly using a conditional random field [35]. Conditional Random Fields are a family of conditionally trained undirected graphical models used to calculate the conditional probability of values on designated output nodes given values assigned to other designated input nodes [35]. The CRF layer in our model was designed to select the best tag sequence from all possible tag sequences by considering the outputs from BiLSTM and correlation between adjacent tags.

## V. Data Preparation

### 1. Annotation Standards

The currently available NER datasets are mostly annotated according to Stanford, IOB1 or IOB2 standards. Although all of these annotation standards aim to mark the tag of NEs, there exists subtle differences among these standards. The details of these standards are shortly introduced below.

### A. Stanford Standard

In the Stanford annotation standard, a tag representing the NE type should be assigned for each token. It does not address the consecutive tokens in the sentence. Figure 4 shows an example Amharic sentence annotated by the Stanford annotation standards.

*Figure 4: Example of Stanford annotation standard*

| | |
|---|---|
| አርቲስቱ (the artist) | O |
| ትናንት (yesterday) | O |
| ለዋልታ (to walta) | ORG |
| እንፎርሜሽን (information) | ORG |
| ማእከል (center) | ORG |
| እንደገለጸው (as expressed) | O |
| የአዋሳ (of Awasa) | LOC |
| ፣ (,) | O |
| አርባ (Arba) | LOC |
| ምንጭ (Minch) | LOC |
| ፣ (,) | O |
| ናዝሬትና (Nazaret) | LOC |
| መቀሌ (Mekele) | LOC |
| ከተሞችን (cities) | O |

### B. Inside-Outside-Beginning 1 (IOB1)

Likewise the Stanford Annotation Standard, IOB1 also assigns a tag for each token [53]. However, unlike the Stanford Annotation Standard, IOB1 tags consist of two parts: a position marker (I, O or B) and the type of NE. The position markers I, O and B represent Inside, Outside and Beginning respectively. An NE, whether be a single-word or multi-word noun phrase, between two non-NEs are denoted by I position marker. B position is only used when the first word of an NE phrase immediately follows another NE phrase. Words marked as O for the non-NE tokens. Figure 5 shows example of the IOB1 annotation standards.

*Figure 5: Example of IOB1 annotation standard*

| | |
|---|---|
| አርቲስቱ (the artist) | O |
| ትናንት (yesterday) | O |
| ለዋልታ (to walta) | I-ORG |
| እንፎርሜሽን (information) | I-ORG |
| ማእከል (center) | I-ORG |
| እንደገለጸው (as expressed) | O |
| የአዋሳ (of Awasa) | I-LOC |
| ፣ (,) | O |
| አርባ (Arba) | I-LOC |
| ምንጭ (Minch) | I-LOC |
| ፣ (,) | O |
| ናዝሬትና (Nazaret) | I-LOC |
| መቀሌ (Mekele) | B-LOC |
| ከተሞችን (cities) | O |

---

³ xxx is NE type

### C. Inside-Outside-Beginning 2 (IOB2)

In IOB2 annotations, a B tag is used to mark for all base noun phrases initial words [56]. Figure 6 shows example of the IOB2 annotation standards.

*Figure 6: Example of IOB2 annotation standard*

| | |
|---|---|
| አርቲስቱ (the artist) | O |
| ትናንት (yesterday) | O |
| ለዋልታ (to walta) | B-ORG |
| እንፎርሜሽን (information) | I-ORG |
| ማእከል (center) | I-ORG |
| እንደገለጸው (as expressed) | O |
| የአዋሳ (of Awasa) | B-LOC |
| ፣ (,) | O |
| አርባ (Arba) | B-LOC |
| ምንጭ (Minch) | I-LOC |
| ፣ (,) | O |
| ናዝሬትና (Nazaret) | B-LOC |
| መቀሌ (Mekele) | B-LOC |
| ከተሞችን (cities) | O |

### 2. Current Amharic NER Dataset

Most of the state-of-the-art studies using a corpus-based approach by applying different machine learning algorithms necessitates large amounts of data. However, the Amharic language is a low-resource language and suffers from the lack of a large annotated corpus. The only available Amharic NER corpus has 13,000 tokens [3], which is not efficient to train deep learning models. Additionally, the annotation scheme of this NER dataset is based on the Stanford standard which causes ambiguous representations in entities constructed by consecutive tokens. Converting this annotation format to IOB format is a non-trivial task for Amharic language because two consecutive person, location or organization name can appear together in a sentence without a separation character such as comma (see Figure 7). IOB2 standard assigns these consecutive names as B-xxx and I-xxx³. However, the correct tag should be B-xxx and B-xxx. Therefore, building this IOB2 style annotated corpus is important and useful for the success and standardization of Amharic NER research. Since named entity recognition is proposed as a word-level tagging problem, all of the proposed data sets use word-level tags to denote named entity phrases. A named entity phrase may span multiple words; hence a NE tag is composed of concatenation of a position indicator (B-Beginning, I- Inside) and a NE type (PER, ORG, LOC, etc.). In addition, an O tag indicates that a token is not inside a NE phrase. A named entity phrase starts with a B- tag and if it consists of multiple words, the following word tags are prefixed with I-. We started our work with defining the NER task for Amharic. We prepared short guidelines for tagging Amharic corpus. The guidelines are prepared based on the unique properties of the Amharic language. Finally, the first annotator labels 182K tokens and 20k tokens randomly selected and re-annotated by the second annotator by using the IOB2 coding scheme (see *Figure 9*). After that, we measured inter-annotator agreement by Cohen's Kappa score on the 20K tokens annotated by multiple annotators. The tagging agreement measured by kappa score and achieved 0.7321. According to Cohen Kappa result analysis, our result is interpreted as a substantial agreement between taggers [42] (see Table 2). The highest difference in tagging is observed on identifying I-ORG.

*Figure 7: Sample consecutive names appear for the sentence:"እህመድና ሰለሞን ወደ ሱቅ እየሄዱ ነው ።"-Ahmed and Solomon are going to a shop.*

| | |
|---|---|
| እህመድና/aäḥmädena/ | B-PER |
| ሰለሞን/sälämon/ | B-PER |
| ወደ/wädä/ | O |
| ሱቅ/suq/ | O |
| እየሄዱ/eyähedu/ | O |
| ነው/näwu/ | O |
| ።/./ | O |

TABLE 2. INTERPRETATION OF COHEN'S KAPPA

| No | Value of Kappa | Level of Agreement |
|---|---|---|
| 1 | 0 | Agreement equivalent to chance |
| 2 | 0.1-0.20 | Slight agreement |
| 3 | 0.21-0.40 | Fair agreement |
| 4 | 0.41-0.60 | Moderate agreement |
| 5 | 0.61-0.80 | Substantial agreement |
| 6 | 0.81-0.99 | Near perfect agreement |
| 7 | 1 | Perfect agreement |

## 3. Data Collection

In this study, we have used an Amharic corpus which is prepared by the Ethiopian Languages Research Center (ELRC) of Addis Ababa University in a project called "The Annotation of Amharic News Documents". The project was to manually tag each Amharic word in its context with the most appropriate parts-of-speech. This corpus is prepared in two forms i.e. in Amharic version (using Ge'ez Fidäl) and in transliterated format (SERA) using Latin characters. However, we have chosen to use both Amharic version and the transcription one separately in order to measure the efficiency of the model on spelling variation in the Amharic script. The corpus has 210,000 words collected from 1,065 Amharic news documents of Walta Information Center, a government news and information service located in Addis Ababa, Ethiopia. The previous researcher [3] tagged only 13,538 tokens based on Stanford notation. These datasets are not enough to build a deep neural network model and the standard used is different from IOB2. Therefore, we decided to prepare our own dataset. From the total Walta Information Center dataset, we tagged 8,070 sentences, which has 182,691 tokens. *Figure 8* shows a sample ELRC document format:

*Figure 8: The ELRC document format*

```
<document>
<filename>mes06a1.htm</filename>
<title>የኢትዮጵያ <NP> የትምህርት <NP> ሹፈኝ <N> በ6 ካተብ 4
<NUMP> በመቶ <NUMP> ማደጉ <VN> ተገለጸ <V> ።
<PUNC></title>
<dateline place="አዲስ አበባ" month="መስከረም"
date="6/1994/(WIC)/" />
<body>
የህገሪቱ <NP> የመጀመሪያ <NP> ደረጃ <N> ትምህርት <N> ሹፈኝ <N>
በ6 <NUMP> ካተብ <N> 4 <NUMCR> በመቶ <NUMP> እድገት
<N> ሲያሳይ <VP> ትምህርቷቸውን <NCR> የሚያቋርጡና <VPC>
የሚደግሙ <VP> ተማሪዎች <N> ቁጥር <N> በሁለት <NUMP> በመቶ
<NUMP> መቀነሱን <VN> የትምህርት <NP> ሚኒስቴር <N> አስታወቀ
<V> ። <PUNC>
<copyright>copyright 1998 - 2002 Walta Information
Center</copyright>
</body>
</document>
```

## 4. Building Dataset

Prior to the annotation process, we conduct a couple of cleaning operations on the corpus. The Walta Information Center Corpus is in XML format. The corpus contains some XML tags (e.g. <document>, <filename>, <title>), which are unnecessary for our purposes. Since we have used the body part of the document, all other parts and tags of the documents are removed.

Manual Tagging on the cleaned dataset is conducted by employing the Inside-Outside-Beginning (IOB2) tagging scheme. Annotators tagged each word/token with other (O) or one of three entity types: Person, Location, or Organization. A single named entity could contain several tokens within a sentence. We use the IOB2 annotation standard for Amharic NER tasks. Figure 9 shows the tagging scheme in the IOB2 of Amharic corpus with an example sentence along with its English translation.

**Amharic:** አርቲስቱ ትናንት ለዋልታ እንፎርሜሽን ማዕከል እንደገለጸው የአዋሳ ፣ አርባ ምንጭ ፣ ናዝሬትና መቀሌ ከተሞቻችን በጽዳት ለማስዋብ ፕሮጀክት ነድፎ በአሁኑ ወቅት ስራ ጀምራል ።

***English:*** *The artist told Walta Information Center yesterday that a project to clean up the towns of Awassa, Arba Minch, Nazareth and Mekelle has been launched.*

**Figure 9: Amharic NER dataset tagging example for a sample sentence**

| | |
|---|---|
| አርቲስቱ (the artist) | O |
| ትናንት (yesterday) | O |
| ለዋልታ (to walta) | B-ORG |
| እንፎርሜሽን (information) | I-ORG |
| ማዕከል (center) | I-OR |
| እንደገለጸው (as expressed) | O |
| የአዋሳ (of Awasa) | B-LOC |
| ፣ (,) | O |
| አርባ (Arba) | B-LOC |
| ምንጭ (Minch) | I-LOC |
| ፣ (,) | O |
| ናዝሬትና (Nazaret) | B-LOC |
| መቀሌ (Mekele) | B-LOC |
| ከተሞችን (cities) | O |
| በጽዳት (by cleaning) | O |
| ለማስዋብ (to beautify) | O |
| ፕሮጀክት (project) | O |
| ነድፎ (design) | O |
| በአሁኑ (in the current) | O |
| ወቅት (time) | O |
| ስራ (work) | O |
| ጀምራል (started) | O |
| ። (.) | O |

Since the corpus is in Amharic alphabet, it may be confusing for those who are unfamiliar with the symbols. To help non-native users, Text Romanization was applied on the raw dataset. Text Romanization is the process of converting the non-Latin letters into Latin letters. Amharic

NLP researchers use the System for Ethiopic Representation in ASCII (SERA) transcription system [23]. SERA transcription system uses the same character for different orthographic (for example, the SERA equivalent of "ህ","ሃ","ኀ", "ሐ", "ሃ","ኀ" and "ኃ" is "ha"). Using transliterated corpus reduced spelling variations, which are common in Amharic script (Fidäl) writing. The spelling variation problem is also common for other NLP tasks like IR systems. We have two varieties of corpus, which are the Amharic script and Latin script. It is also possible to reduce the spelling variation by using normalization.

*Table 3: Proportion of Amharic NER dataset*

| **Entity Type** | **Token Count** | **Token Percent** |
|---|---|---|
| Person | 3,809 | 2.08% |
| Location | 7,199 | 3.94% |
| Organization | 7,596 | 4.16% |
| O | 164,087 | 89.82% |
| Total | 182,691 | |

As *Table 3* indicates, our dataset classes are imbalanced. A feed-forward neural network trained on an imbalanced dataset may not learn to discriminate enough between classes [15]. They calculated an attention factor from the proportion of samples presented to the neural network for training. The learning rate of the network elements was adjusted based on the attention factor. In case of imbalanced datasets, the classifier mostly classifies entities as a majority class member [2]. Since the dataset is imbalanced, we applied SMOTE like Belay [8] and Demissie [14] on the dataset and tested our model based on the new datasets.

SMOTE is a means of increasing the sensitivity of a classifier to the minority class by balancing the number of instances between the classes. *Table 4* shows the number of instances for each class after applying SMOTE.

*Table 4: Amharic NER datasets after applying SMOTE*

| **Entity Type** | **Token Count** | **Token Percent** |
|---|---|---|
| Person | 107,744 | 20.00% |
| Location | 153,745 | 28.50% |
| Organization | 149,584 | 27.75% |
| O | 127,980 | 23.75% |
| Total | 539,053 | 100.00% |

## VI. Experimental Results and Evaluation

### 1. Experimental Setup

The hyper parameters we used were proposed by Lample et al. [36]. The researcher experimented with the different hyper parameters and proposed the best ones. In addition to this we have also experimented for epoch size and found a maximum of 50. We used the size of word embeddings as 300, the size of the character embeddings as 25, and the batch size as 20. We use Adam [69] with the learning rate of 0.001 for optimization. Adam is an optimization algorithm that can be used instead of the stochastic gradient descent procedure to update network weights iteratively based on training data. To mitigate overfitting, we apply the dropout method to regularize our model by applying dropout on both the input and output vectors of BiLSTM. We fix the dropout rate at 0.5 for all dropout layers through all the experiments.

### 2. Evaluation Metrics

There are different types of named entity evaluation metrics. These are MUC, CoNLL, SemEval and others.

#### A. CoNLL Metric

The Language-Independent Named Entity Recognition task introduced at CoNLL-2003 [57] measures the performance of the systems in terms of precision, recall and $F\_1$ score, where: Precision is the percentage of named entities found by the learning system that are correct. Recall is the percentage of named entities present in the corpus that are found by the system. A named entity is correct only if it is an exact match of the corresponding entity in the data file [65].

$$Precision = \frac{True\ Positive}{True\ Positive + False\ Positive} \quad (6)$$

$$Recall = \frac{True\ Positive}{True\ Positive + False\ Negative} \quad (7)$$

$$F\_1 = \frac{2\ (Precision * Recall)}{Precision + Recall} \quad (8)$$

### B. MUC (Message Understanding Conference) Metric

MUC considered different categories of errors. It defined in terms of comparing the response of a system against the golden annotation [10]:

**Correct (COR):** both are the same
**Incorrect (INC):** the output of a system and the golden annotation do not match
**Partial (PAR):** system and the golden annotation are somewhat "similar" but not the same
**Missing (MIS):** a golden annotation is not captured by a system
**Spurious (SPU):** system produces a response, which doesn't exist in the golden annotation

**Possible** contain the tallies of the number of slot fillers that should be generated. It is the sum of the correct, partial, incorrect, and missing. **Actual** is the number of fillers that the system under evaluation actually generated, which is the sum of the correct, partial, incorrect, and spurious. Based on this they calculated Precision, Recall and $F\_1$ score;

$$Recall = \frac{correct + (0.5 * partial)}{Possible} \quad (9)$$

$$Precision = \frac{Correct + (0.5 * partial)}{Actual} \quad (10)$$

$$F\_1 = \frac{2\ (Precision * Recall)}{Precision + Recall} \quad (11)$$

### C. 6.2.3 International Workshop on Semantic Evaluation (SemEval)

The evaluation metrics should score if a system is able to identify the exact entity (regardless of type) and it is able to assign the correct entity type regardless of the boundaries [59].
The SemEval'13 introduced four different ways to measure precision, recall and $F\_1$ score results based on the metrics defined by MUC.

**Strict**: exact boundary surface string match and entity type
**Exact**: exact boundary match over the surface string, regardless of the type
**Partial**: partial boundary match over the surface string, regardless of the type
**Type**: some overlap between the system tagged entity and the gold annotation is required

In order to calculate precision and recall, it used the scoring categories proposed by MUC (COR, INC, PAR, MIS, and SPU) in different ways.

For both the boundaries and the type, the following measure are calculated:

**COR**: The number of correct answers

The number of annotations in the gold standard, which contribute to the final score:

$$Possible\ (POS) = COR + INC + PAR + MIS = TP + FN \quad (12)$$

The total number of annotations produced by the system:

$$Actual\ (ACT) = COR + INC + PAR + SPU = TP + FP \quad (13)$$

After that, it computed precision. Recall and $F_1$ score. The computation is made in two different ways depending on whether we want an exact match (i.e., strict and exact) or a partial match (i.e., partial and type) scenario.

Exact Match (i.e., strict and exact):

$$Precision = \frac{COR}{ACT} = \frac{TP}{TP+FP} \quad (14)$$

$$Recall = \frac{COR}{POS} = \frac{TP}{TP+FN} \quad (15)$$

Partial Match (i.e., Partial and Type)

$$Precision = \frac{COR+(0.5*PAR)}{ACT} = \frac{TP}{TP+FP} \quad (16)$$

$$Recall = \frac{COR+(0.5*PAR)}{POS} = \frac{COR}{ACT} = \frac{TP}{TP+FP} \quad (17)$$

We have used CoNLL evaluation metrics of precision, recall and $F_1$ score [65] for our experiments. Among the evaluation metrics defined above, the CoNLL metric is the harshest one. The CoNLL evaluation metric is an aggressive metric where partially matched NE tokens cannot contribute to the overall score. An NE has to be identified as a whole and its type must be correctly classified in order to gain credit. So, in order to evaluate our system in an aggressive manner, we opt to use the CoNLL metric in our evaluations.

### 3. Experiments

We have compared different models such as BiLSTM, BiLSTM-CRF and RoBERTa and tested them using our dataset and SAY dataset. The size of the SAY dataset is almost half of our dataset. Details of the datasets are presented in the dataset section. For these models we used two-third of the dataset for training and one-third of the dataset for testing. In order to develop a pre-training language model we have collected datasets from different sources like: news, tweeter, and web. The dataset has more than 6 million sentences. Based on this dataset we have developed fastText and RoBERTa language models.

To compare the two datasets using the BiLSTM-CRF model, we change the Title and Time classes to O because our dataset does not have these classes. *Table 5* shows that increasing the size of the dataset increases the performance of the system. The experimental result shows that there is an increase of 0.47 % of F1-score from BiLSTM to BiLSTM-CRF. In this experiment we have also compared our fastText word vector with Facebook fastText for the BiLSTM-CRF model. The result shows that our new fastText word vector is better than Facebook fatsText. This shows that increasing the size of word vectors improves the performance of the system. In addition to this, we have also compared pre-defined word vectors with the transformer learning (Roberta) model. As we can see from *Table 5* RoBERTa is much better than fastText embeddings for Amharic. The RoBERTa model improves the Amharic NER system by 1.19%.

*Table 5: Comparison of tagging performance in Precision, Recall and F_1 scores*

| | **Model** | **Dataset Type** | **Precision** | **Recall** | **F_1 Score** |
|---|---|---|---|---|---|
| Pre-defined Word Vectors (Facebook) | BiLSTM-CRF | SAY | 57.99 | 70.32 | 63.56 |
| Pre-defined Word vectors (own) | BiLSTM-CRF | SAY | 60.76 | 67.18 | 63.81 |
| Pre-defined Word Vectors (Ours) | BiLSTM | Ours | 66.98 | 77.70 | 71.94 |
| Pre-defined Word Vectors (Facebook) (Own dataset) BiLSTM-CRF | BiLSTM-CRF | Ours | 68.71 | 76.28 | 72.30 |
| Pre-defined Word Vectors (own) | BiLSTM-CRF | Ours | 70.10 | 74.87 | 72.41 |
| Amharic RoBERTa | RoBERTa | Ours | 71.44 | 75.87 | 73.60 |

We measured the performance of our model by using different hyper-parameters: randomly initialized word vector, pre-defined word vector (fastText), using SMOTE experiments. For these experiments, we have used 10-fold cross validation for the original dataset and for SMOTE experiments 80% for training and the remaining 20% for testing. For each experiment, we have tested the model and presented it in *Table 6*.

In the first experiment, the character based BiLSTM-CRF model based on random initial word vectors was used. The random initial word vector uses the random initial value for the word vector through the training process. Taking character embedding 25 and word embedding size of 300. This Amharic NER system achieved Precision, Recall and F_1 score values of 66.41%, 74.72% and 70.18% respectively as shown in *Table 6*.

In the second experiment, a character-based BiLSTM-CRF model was built based on predefined word vectors (fastText). Language model pre-training has been shown to be effective for improving many natural language processing tasks [17] [50] [32] [16]. These include sentence-level tasks such as natural language inference and paraphrasing, which aim to predict the relationships between sentences by analysing them at once, as well as token-level tasks such as named entity recognition and question answering, where models are required to produce fine-grained output at the token level [65] [52]. The pre-trained word vector is generated from large datasets collected from the different sources. Each word in the corpus has its own vector values. The fastText word vectors of Amharic were trained using CBOW with position weights in dimension 300, with character gram of length 5, and 10 negative. This model achieved a better performance compared with the first model. The model achieved 72.92, 75.37, and 74.12 for precision, recall and F_1 score respectively. Incorporating pre-trained word vectors increased the performance by 3.94. With a deeper analysis per each fold, we noticed that pre-trained word vectors consistently outperform the randomly initialized vectors. That suggests, pre-trained word vectors can consistently achieve better performance than the randomly initialized word vectors even though the confidence intervals overlap.

In order to see the effects of SMOTE to alleviate imbalanced data problems, we prepared two setups. In the first setup, we split our original imbalanced dataset in 80% training and 20% test and applied SMOTE only on training data.

By this way, we aim to balance the class distribution in the training dataset to help our model learn better whereas we keep the original distribution of classes in the test dataset, which is the actual case in real-world. The second setup is prepared by applying SMOTE both on training and test splits to determine the upper bound of the model if the classes were distributed equally. We didn't implement cross-validation in these setups since applying SMOTE technique considerably increases the training dataset size (from 182K to 539K). In the experiments on these two setups, we used the fastText predefined word vectors because they were proven to be best performing in our previous experiments. The results of these experiments are shown in the last two rows of *Table 6*. We achieved the state-of-the-art results for Amharic NER in our experiments.

*Table 6: Experimental results in Precision, Recall and F_1 scores*

|  | **Precision** | **Recall** | **F_1 Score** |
|---|---|---|---|
| Random Initialized Word Vectors | 66.41±4.65 | 74.72± 4.12 | 70.18± 3.67 |
| Pre-defined Word Vectors (fastText) | 72.92 ±2.75 | 75.37 ±4.00 | 74.12 ±3.50 |
| Applying SMOTE Only for Training Datasets | 91.42 | 95.01 | 93.18 |
| SMOTE for All Datasets | 98.75 | 99.15 | 98.95 |

We also conduct an analysis to measure the effect of the dataset size on the NER performance. We trained our best-performing setup with datasets of different sizes and measured the performance. The results are depicted in the learning curve presented in *Figure 10*. As expected, the more data we have, the better F_1 scores we get. The positive incline on the right-hand side of the learning curve represents we have not converged yet, in other terms, even higher F_1 score may be achieved with more training dataset.

---

[4] https://github.com/Ebrahimc/ANEC-An-Amharic-Named-Entity-Corpus-

The Amharic fastText word vectors have 304,649 words, increasing the size of the word vector decreases out of vocabulary and increases the performance of the model [37]. Similar to Belay [8] and Demissie [14], using balanced classes in the dataset increased the performance of Amharic NER system. The final version of our Amharic NER Dataset is made public[4].

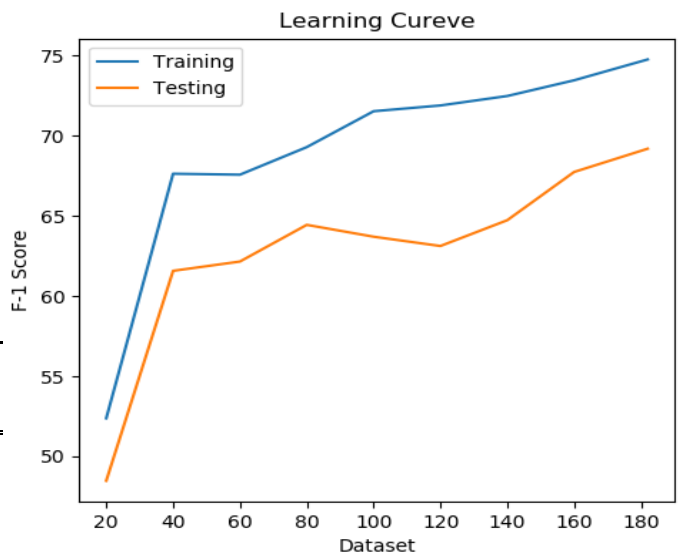

**Figure 10: Learning curve shows the F_1 score over size of dataset**

## VII. Conclusions and Further Work

In this study, we built a publicly available and relatively large Amharic NER dataset and achieved the state-of-the-art NER results by using BiLSTM-CRF deep learning models.

One of the obstacles in improving the Amharic NER studies was the lack of high quality and publicly available NER datasets. We address this issue in our study by manually tagging a relatively large Amharic NER dataset which is composed of 8.070 sentences and 182K tokens taken from Walta Information Center Dataset. To assess the quality of the annotation process, we also employ inter-annotation agreement measurements where we got a 0.7321 Kappa score denoting there is a substantial agreement between annotators. The final version of this dataset is made publicly available to serve as a baseline for further NER and other NLP-related studies on Amharic. Also, we would like to extend our efforts to increase the size of this corpus as well as to increase the variety of the corpus by acquiring textual content from sources other than the Walta Information Center Dataset.

Another contribution of our study is an Amharic NER system developed by deep learning models employing a CRF classifier on top of a BiLSTM layer. One of the findings from our experimental studies is that utilizing pre-trained word vectors trained on a comparatively large corpus performs better instead of randomly initialized word vectors. To overcome the problem of imbalanced class distribution of the dataset, we evaluate the effects of the SMOTE oversampling method. Our best performing configuration achieves to get a new Amharic NER state-of-the-art score of 93.18% as F_1 score, which is significantly outperforming the baseline.

Recent studies on the NER task for other languages take the advantage of transfer learning by using transformer architectures like BERT [16]. Transformer architectures need to optimize a large number of weights, which necessitates huge amounts of text. We have collected around 6 million dataset from news, tweeter and web corpus for developing the RoBERTa model. The RoBERTa model improves the Amharic NER system by 1.19%.

Having morphological features as input is shown to be beneficial in many of the previous studies. Although an Amharic morphological analyser is available [25], we did not augment our word vectors with morphological features since no morphological disambiguation tool is accessible. Since the outputs of the morphological analysis phase are ambiguous, morphological disambiguation is inevitable for Amharic. To improve our NER results and support future NLP-related studies in Amharic, we also started to work on the Amharic morphological disambiguation task. We expect to improve our current state-of-the-art Amharic NER result with the contribution of these morphological features.

## Acknowledgments

At this work, Ebrahim Chekol Jibril was supported by the Turkcell - Istanbul Technical University Researcher Funding Program.

## Appendix I: The algorithm of SMOTE

**Algorithms SMOTE (T, N. k)**

**Input**: Number of minority class sample T; Amount of SMOTE N%; Number of nearest     neighbors k

**Output**: (N/100) * T synthetic minority class samples

   (* if N is less than 100%, randomize the minority class sample as only a random     percent of them will be SMOTE. *)

**If** N < 100

   **then** Randomize the T minority class samples

         T = (N/100)*T ,          N = 100

   **endif**

N = (int)(N/100) (* The amount of SMOTE is assumed to be in integral multiple of 100. *)

k = Number of nearest neighbors,

numattrs = Number of attributors

sample [][] : array for original minority class samples

newindex: keeps a count of number of synthetic samples generated, initialized to 0

Synthetic[][]: array for synthetic samples (* Compute k nearest neighbors for each minority class sample only*)

**for** i <---1 to T

   Compute k nearest neighbors for i, and save the indices in the nnarray

   Populate(N, i, nnarray)

**endfor**

Populate(N, i, nnarray)(* Function to generate the synthetic samples. *)

**while** N !=0

   Choose a random number between 1 and l, call it nn. This step schooses one of the k nearest neighbors of i.

   **for** attr <-- 1 to numattrs

      Compute: dif = Sample[nnarray[nn]][attr]-Sample[i][attr],     Compute: gap = random number between 0 and 1

      Synthetic[newindex][attr] = Sample[i][attr] + gap*dif

   **endfor**

   newindex++,

   N = N - 1

**endwhile**

**return** (*End of Populate. *)

End of Pseudo-Code.

## Appendix II: Amharic NER Annotation Guideline

We have prepared annotation guideline for Person, Location and Organization entities.

**Person**

When the entity refers to an individual or collective person (more than one individual) including fictitious persons. Even in the case of a collective person annotation, there must be the presence of a proper name.Titles/prefixes such as "አቶ/Mr.", ወ/ሮ/Mrs, ፕሮፌሰር/prof, ዶክተር/Dr., አርቲስት/Artist, etc. and role names such as "President" are not considered part of a person name.

"ዶክተር አህመድ"
ዶክተር <B_ENAMEX TYPE="PERSON">አህመድ፡ <E_ENAMEX>
"ሴክሬታራ ብሌን ስዩም"
ሴክሬታሪ <B_ENAMEX TYPE="PERSON">ብሌን ስዩም"<E_ENAMEX>
"ሚኒስቴር አለበል"
ሚኒስቴር <B_ENAMEX TYPE= "PERSON"> አለበል <E_ENAMEX>
Family names are to be tagged as PERSON.
"የበላይ ቤተሰብ" the <B_ENAMEX TYPE="PERSON">"የበላይ <E_ENAMEX> ቤተሰብ

Although religious titles or specifiers such as "ቅዱስ/saint," "ነብይ/prophet," and "ኢማም/imam," are not be tagged, the proper name will be tagged as a PERSON. This practice becomes more significant in marking up speech transcriptions, due to peculiarities of speech habits or patterns.

ቅዱስ ፓውሎስ ወደ ከተማ ገቡ
ቅዱስ <B_ENAMEX TYPE="PERSON">ፓውሎስ <E_ENAMEX> ወደ ከተማ ገቡ
"ኢማም አህመድ ወደ መስጊድ ሄዱ."
ኢማም <B_ENAMEX TYPE="PERSON">አህመድ፡ <E_ENAMEX> ወደ መስጊድ ሄዱ

Names of fictional characters are to be tagged; however, character names used as TV show titles will not be tagged when they refer to the show, rather than the character name.

አስፓይደርማን የልጆች ፊልም ጥሩ አክተር ነዉ
<B_ENAMEX TYPE="PERSON">አስፓይደርማን <E_ENAMEX> የልጆች ፊልም ጥሩ አክተር ነዉ

**Organizations**

A company which sells products or provides services that are not only administrative. It includes both private and public companies, as well as hospitals, schools, universities, political parties, trade unions, police, gendarmerie, mosques, churches, sportive clubs, etc.
An organization which plays a mainly administrative role. It is often an administrative and/or geographical division. This includes town halls, city council, regional council, state council, federal council, named government, ministry parliament, prefectures, ministries dioceses, tribunal, court, government treasury, public treasury, international org.

Corporate or organization designators such as ኮርፖሬሽን/Co." are considered part of an organization name.
"አፐል ኮርፖሬሽን "
<B_ENAMEX TYPE="ORGANIZATION">አፐል ኮርፖሬሽን <E_ENAMEX>
Proper names that are to be tagged as ORGANIZATION include stock exchanges, multinational organizations, businesses, TV or radio stations, political parties, religions or religious groups, orchestras, bands, or musical groups, unions, non-generic governmental entity names such as "Congress" or "Chamber of Deputies," sports teams and armies (unless designated only by country names, which are tagged as LOCATION), as well as fictional organizations (to ensure consistency with marking other fictional entities).
"የኢትዮጵያን ኮሚኒቲ"
<B_ENAMEX TYPE="ORGANIZATION">የኢትዮጵያን ኮሚኒቲ<E_ENAMEX>
"የኢትዮጵያን ቡና ክለብ"
<B_ENAMEX TYPE="ORGANIZATION">የኢትዮጵያን ቡና ክለብ<E_ENAMEX>
"ዛሬ በኢትዮጵያን ቴሌቪዥን"
ዛሬ<B_ENAMEX TYPE="ORGANIZATION">በኢትዮጵያን ቴሌቪዥን<E_ENAMEX>

Stations, channels, and call numbers are to be tagged as ORGANIZATION; however, when a location is given which is the location of the radio station transmitter, the location will be tagged separately:

"አዲስ ኤፍ ኤም 97.1"

Although event names are not to be tagged, even if they refer to events that occur on a regular basis and are associated with institutional structures, the institutional structures themselves -- steering committees, etc. – should be tagged as ORGANIZATION.

"የኢትዮጵያ አትሌቲክስ ኮሚቴ"

<B_ENAMEX TYPE="ORGANIZATION">የኢትዮጵያ አትሌቲክስ ኮሚቴ<E_ENAMEX>

"ሸራተን አዲስ ሆቴል"

<B_ENAMEX TYPE="ORGANIZATION">ሸራተን አዲስ ሆቴል<E_ENAMEX>

"አንሞር መስጊድ እና ኡራኤል ቢተክርስቲያን ጎን ለጎን ናቸው::"

<B_ENAMEX  TYPE="ORGANIZATION">  አንሞር  መስጊድ  <E_ENAMEX>  እና  <B_ENAMEX TYPE="ORGANIZATION">  ኡራኤል ቢተክርስቲያን <E_ENAMEX> ጎን ለጎን ናቸው::

"በዚህ ሳምንት አዲስ ዘመን ያወጣዉ ዘገባ..."

በዚህ ሳምንት <B_ENAMEX TYPE= "ORGANIZATION"> አዲስ ዘመን <E_ENAMEX> ያወጣዉ ዘገባ...

**Locations**

The TYPE LOCATION applies to entities representing either geographical, political, or astronomical locations. Examples of strings that are tagged as LOCATION include: continents, countries, provinces, counties, cities, regions, districts, towns, villages, neighbourhoods, airports, military bases, railways, railroads, highways, bridges, street names, street addresses, oceans, seas, straits, bays, channels, sounds, rivers, islands, lakes, national parks, mountains, fictional or mythical locations.

The following are examples of locations:

"ኤድናሞል ሲኒማ"

<B_ENAMEX TYPE="LOCATION">ኤድናሞል <E_ENAMEX>

"አዲስ አበባ ኢተርናሽናል ኤርፖርት ..."

<B_ENAMEX TYPE="LOCATION"> አዲስ አበባ ኢተርናሽናል ኤርፖርት <E_ENAMEX>...

- Compound expressions in which place names are listed in succession, with or without a separating comma, are to be tagged as separate instances of LOCATION.

  "አዲስ አበባ ፣ ባህር ዳር እና ጅማ ..."

  <B_ENAMEX TYPE="LOCATION"> አዲስ አበባ <E_ENAMEX>፣ <B_ENAMEX TYPE="LOCATION">ባህር ዳር <E_ENAMEX>እና <B_ENAMEX TYPE="LOCATION">ጅማ <E_ENAMEX>...

- Designators that are integrally associated with a place name are to be tagged as part of the name. For example, include in the tagged string the word "*ወንዝ*/River" in the name of a river, "*ተራራ*/Mountain" in the name of a mountain, "*ከተማ*/City" in the name of a city, etc., if such words are contained in the string.

  "አባይ ወንዝ መነሻዉ ..."

  <B_ENAMEX TYPE="LOCATION">አባይ ወንዝ<E_ENAMEX>...

- A place name can be included in the name of organization

  "አዲስ አበባ ዩኒቨርስቲ ..."

  <B_ENAMEX TYPE=" ORGANIZATION ">አዲስ አበባ ዩኒቨርስቲ <E_ENAMEX>...

- Street/ጎዳና names will be tagged as a single expression that includes the noun Street.

  አትሌት ኃይለጉብር ስላሴ ጎዳና

  አትሌት <B_ENAMEX TYPE="LOCATION">ኃይለጉብር ስላሴ ጎዳና<E_ENAMEX>...

- Directional expressions(north/ሰሜን, south/ደቡብ, etc. ) will be tagged only when they are part of the official name

  ምስራቅ ጎጃም የሀዝቡ ለስራ ያለዉ ተነሳሽነት ...

  <B_ENAMEX TYPE="LOCATION"> ምስራቅ ጎጃም <E_ENAMEX>የሀዝቡ ለስራ ያለዉ ተነሳሽነት ...

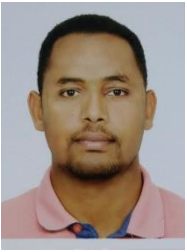

**Ebrahim Chekol Jibril** received the B.Sc. degree in Information Technology and M.Sc. degree in Information Science from Addis Ababa University, Ethiopia in 2005 and 2010 respectively. He is currently pursuing Ph.D. degree in computer engineering at Istanbul Technical University. He has worked as a senior lecturer in the faculty of Computing, Bahir Dar Institute of Technology, Bahir Dar University. His research interests include natural language processing, machine learning and deep neural network.

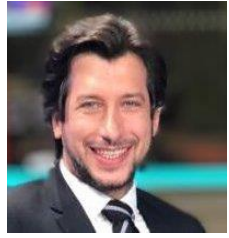

**A.CÜNEYD TANTUĞ** received the master's and PhD degrees from the Istanbul Technical University, Turkey. He is currently an Associate Professor in the Faculty of Computer and Informatics at Istanbul Technical University. His research interests include natural language processing, machine translation and deep learning.